%% file: acl_latex.tex
\documentclass[11pt]{article}

\usepackage[final]{acl}

\usepackage{times}
\usepackage{latexsym}
\usepackage{amssymb}
\usepackage{tikz}
\usepackage[T1]{fontenc}

\usepackage[utf8]{inputenc}

\usepackage{microtype}
\usepackage{tikz}
\newcommand{\numcircledtikz}[1]{%
  \tikz[baseline=(char.base)]{
    \node[shape=circle,draw,inner sep=1pt,font=\small] (char) {#1};
  }%
}
\usepackage{inconsolata}

\usepackage{graphicx}

\usepackage{booktabs}
\usepackage{amsmath}

\title{From 2:4 to 8:16 sparsity patterns in LLMs for Outliers and Weights with Variance Correction}

\author{
 \textbf{Egor Maximov\textsuperscript{1,2,\,$\clubsuit$}},
 \textbf{Yulia Kuzkina\textsuperscript{1,2,\,$\clubsuit$}},
 \textbf{Azamat Kanametov\textsuperscript{1}},
 \textbf{Alexander Prutko\textsuperscript{2}},
\\
 \textbf{Maxim Zhelnin\textsuperscript{3}},
 \textbf{Egor Shvetsov\textsuperscript{4}},
 \textbf{Aleksei Goncharov\textsuperscript{1,2}}
\\
\\
    \textsuperscript{1}Moscow Independent Research Institute of Artificial Intelligence \\
    \textsuperscript{2}MIL Team 
    \textsuperscript{3}MWS AI
    \textsuperscript{4}Applied AI \\
    \textbf{Correspondence:} \href{mailto:egor.maximov@mil-team.com}{egor.maximov@mil-team.com} \\
    \small{$\clubsuit$ indicates equal contribution.}
}

\begin{document}
\maketitle
\begin{abstract}
    As large language models (LLMs) grow in size, efficient compression techniques like quantization and sparsification are critical. While quantization maintains performance with reduced precision, structured sparsity methods, such as N:M sparsification, often fall short due to limited flexibility and sensitivity to outlier weights. We explore 8:16 semi-structured sparsity, demonstrating its ability to surpass the Performance Threshold---where a compressed model matches the accuracy of its uncompressed or smaller counterpart under equivalent memory constraints. Compared to 2:4 sparsity, 8:16 offers greater flexibility with minimal storage overhead (0.875 vs. 0.75 bits/element). We also apply sparse structured patterns for salient weights, showing that structured sparsity for outliers is competitive with unstructured approaches, leading to equivalent or better results. Finally, we demonstrate that simple techniques such as variance correction and SmoothQuant-like weight equalization improve sparse models performance.
\end{abstract}

\section{Introduction}
\label{sec:introduction}

Large language models (LLMs) are rapidly finding a wide range of applications~\cite{veeramachaneni2025large}. However, as these models continue to scale in size and complexity, significant attention has been devoted to studying model compression techniques, particularly quantization and semi-structured sparsification~\cite{frantar2025compression}. The \textbf{Performance Threshold} is defined as the point at which a compressed model maintains comparable accuracy to its uncompressed or smaller counterpart, with equavalent memory usage~\cite{frantar2025compression}. Quantized models often surpass this threshold by effectively reducing the precision of weights while maintaining or even enhancing performance~\cite{li2023model, frantar2025compression}. In contrast, structured sparsity methods, such as N:M sparsification, typically struggle to meet this threshold for several reasons:

\begin{itemize}
    \item \textbf{Limited Flexibility:} In the N:M scheme, N out of M elements in each block must be zeroed, which can restrict the model's adaptability~\cite{zhang2022learning}.
    \item \textbf{Lack of Hardware Support:} Only 2:4 structured sparsity is natively supported on mainstream GPUs, narrowing research to this specific pattern~\cite{hu2024accelerating}.
    \item \textbf{Potential Outlier Removal:} Removing or perturbing even a few outlier weights can disproportionately degrade model accuracy~\cite{dettmers2023spqr}.
\end{itemize}

Consequently, recent compression efforts increasingly favor quantization for its reliability, leaving sparsification methods underdeveloped~\cite{harma2024effective, frantar2025compression}. This gap motivates our investigation into \textbf{8:16 semi-structured sparsity}. We aim to determine whether 8:16 sparsity can surpass the Performance Threshold and demonstrate its viability for real-world applications. While technical implementation details are beyond this work’s scope, our analysis focuses on performance to pave the road for the future research.

\paragraph{Structured Sparse Outliers}


To mitigate sparsification-induced degradation, methods like SPQR restore salient weights via an auxiliary sparse matrix (e.g., CSR)~\cite{dettmers2023spqr}. While salient weights typically constitute only 1–5\% of total weights, unstructured sparsity introduces inefficiencies: irregular memory access limits cache utilization, and metadata overhead grows linearly with non-zeros. In contrast, structured sparsity patterns (e.g., 4:256, 8:256, 16:256) enable predictable memory access and reduced metadata, improving hardware efficiency~\cite{schulte2023best}. These patterns correspond to sparsity levels of 1.5\%, 3.1\%, and 6.25\%, respectively, balancing flexibility with practical constraints. In this work we aim to answer weather structured sparsity formats are enough to store outliers.

\textbf{Our contributions:}

(1) \textbf{We empirically show that the 8:16 sparsity pattern improves model performance.} Specifically, a sparse LLaMa-2-13B model using this pattern achieves the same performance as the dense LLaMa-2-7B model This suggests promising implications for future hardware development with 8:16 sparsity patterns support. 

(2) \textbf{Structured Sparsity for Salient Weights (\textsc{SSP for SW}):}
    We propose and analyze \textsc{SSP for SW}, a structured sparsity patterns for critical weight subsets. This approach, in our experiments, not only enhances computational efficiency but also enables models with outliers stored in structured patterns (\textsc{SSP}) to outperform those using unstructured sparsity for salient weights.   

(3) We propose two techniques for pre and post processing for weights sparsification.  \textit{SmoothQuant-Inspired Rebalancing:} We pre-process weights and activations before sparsification, adapting the SmoothQuant philosophy to balance distributions and improve robustness.   \textit{Variance Correction:} We introduce post-pruning variance adjustments to mitigate performance degradation caused by weight removal—a simple yet effective technique absent in existing literature\footnote{To the best of our knowledge.}.


\section{Sparsity Pattern Selection and Hardware Implications}
\label{sec:pattern_selection}

When designing sparsity patterns for practical deployment, we must balance performance with hardware implementation constraints. While numerous sparsity patterns exist theoretically, hardware-efficient patterns must satisfy several requirements:

\begin{itemize}
    \item \textbf{Memory Alignment}: Patterns should align with hardware memory boundaries (typically powers of two)
    \item \textbf{Metadata Overhead}: Pattern selection bits must be efficiently packable
    \item \textbf{Circuit Complexity}: Decoding logic should be implementable with reasonable silicon area
\end{itemize}

Table~\ref{tab:pattern_comparison} compares key characteristics of common N:M patterns. The 8:16 pattern offers an optimal balance between flexibility and hardware efficiency. While 16:32 achieves marginally better perplexity, its hardware implementation would require significantly more complex circuitry with diminishing performance returns.

\begin{table}[h]
\centering
\resizebox{0.5\textwidth}{!}{%

\begin{tabular}{lcccc}
\toprule
Pattern & Configurations & Bits/Element & PPL RIA & PPL RIA+VC \\
\midrule
2:4 & 6 & 0.75 & 22.526& 16.66 \\
4:8 & 70 & 0.81 & 12.80 & 11.58 \\
\textbf{8:16} & \textbf{12,870} & \textbf{0.88} & \textbf{10.64} & \textbf{9.95} \\
16:32 & 601,080,390 & 1.00 & 9.98 & 9.51 \\
\bottomrule
\end{tabular}
}
\caption{\textbf{Hardware considerations and performance.} Comparison of N:M sparsity patterns for LLaMA3-8B model.  Perplexity (PPL) measured on WikiText-2 using RIA and RIA+VC.}
\label{tab:pattern_comparison}
\end{table}

From a hardware perspective, both 2:4 and 8:16 patterns provide a 50\% theoretical FLOPs reduction and 2$\times$ reduction in memory bandwidth requirements. While empirical measurements for 8:16 acceleration aren't yet possible (as no hardware currently implements this pattern), we can project its performance based on the established relationship between memory bandwidth reduction and inference acceleration in sparse architectures. The existing 2:4 pattern achieves from $\sim$1.5--2$\times$ inference acceleration scaling with matrix size, and we expect similar scaling for 8:16 when implemented in silicon.

The selection of 8:16 over other patterns is deliberate. While 16:32 shows marginally better perplexity (Table~\ref{tab:pattern_comparison}), the main jump occurs from 4:8 to 8:16 yielding a significantly larger improvement. Moreover, hardware implementation of 16:32 would require substantially more complex circuitry for minimal performance gain, making 8:16 the pragmatic optimum for future hardware development.

\section{Related work}
\label{sec:relatedwork}

\textbf{Sparsity Patterns:}  \cite{hu2024accelerating} introduced a new 2:4 (50\%) sparsity pattern for maintaining accuracy while achieving 2x math throughput for GEMM-like operations.  \cite{zhao2024beyond} presents a V:N:M sparsity pattern that achieves higher sparsity and faster computation than standard N:M patterns. Since widely available hardware supports only 2:4 patterns, most of recent research mainly studied this pattern.

\textbf{Sparsification:} Identifying weights importance\protect\footnote{In our work, we use the terms \textbf{salient weights} and weight \textbf{outliers} interchangeably for very important weights.} is a central challenge in weight pruning. Several recent methods have been proposed for identifying such weights in LLMs, including SparseGPT~\cite{frantar2023sparsegpt}, Wanda~\cite{sun2023simple}, and OWL~\cite{yin2023outlier}. Further RIA~\cite{zhang2024plug} improves upon earlier approaches by evaluating the relative importance of each weight within its row and column before pruning.

\textbf{Outliers:} \citeauthor{dettmers2022gpt3} demonstrated that a small number of activation outliers can significantly impact LLM performance, highlighting the link between activation patterns and salient weights. Many subsequent post-training quantization methods adopted similar or identical metrics to identify these key weights~\cite{dettmers2023spqr, xiao2023smoothquant, lee2024owq}. In addition, many approaches were developed to avoid distortions in these outliers during quantization or sparsification by isolating them. For example, \citeauthor{zhelnin2024gift} indentifies columns with most outliers and does not quantize these columns,  \citeauthor{dettmers2023spqr} stores identified salient weights in a separate sparse matrix, applies sparse matrix multiplication, and quantizes the remaining weights with minimal performance degradation. Following these studies we evaluate feasibility of storing outliers in semi-strucutred N:M patterns.

\textbf{Post-Compression Fine-Tuning:} Weight pruning or quantization typically leads to some model degradation~\cite{kharinaev2025investigating} even if salient weights were isolated~\cite{zhelnin2024gift}. Therefore, fine-tuning is often applied afterward. However, full-model fine-tuning is computationally prohibitive and  efficient fine-tuning methods have emerged, these methods can be categorized as follows:
(1)~\textbf{Adapter-based} methods like QLoRA~\cite{dettmers2023qlora}, where low-rank adapters are trained.
(2)~\textbf{Selective-based} methods like GIFT-SW~\cite{zhelnin2024gift}, where only a small subset of weights is fine-tuned. (3)~\textbf{Block-based} methods like EBFT~\cite{guo2024ebft}, where the model is split into independent blocks, each fine-tuned separately.

\section{Methods}
\label{sec:methodology}
In our preliminary experiments\footnote{Preliminary experiments excluded due to space constraints; full ablation study available in supplementary materials.}, we identified the most effective sparsification strategy through an ablation study. Our proposed \textbf{general pipeline} consists of four key stages:

\begin{enumerate}
    \item \textbf{Weights Equalization}: Apply channel-wise scaling $S(W) \rightarrow W_{\text{ec}}$ using to equalize weight magnitudes, as described in Equation~\ref{eq:sq}.
    \item \textbf{Importance-Aware Pruning}: Employ the RIA method on scaled weights to identify and preserve salient weights $W_{\text{salient}}$ (1.5\%--6.2\% of total weights) using high compression structured sparsity patterns $[4{:}256, 8{:}256, 16{:}256]$. Non-salient weights $W_{\neg\text{salient}}$ are pruned using  2:4 or 8:16 sparsity patterns.
    \item \textbf{Variance Correction}: Compensate for distribution shifts in pruned weights through variance-preserving rescaling of $W_{\neg\text{salient}}$ as defined in Equation~\ref{eq:vc}.
    \item \textbf{Blockwise Fine-Tuning}: Perform error-bound aware tuning using EBFT~\cite{guo2024ebft}, updating only $W_{\neg\text{salient}}$ and BatchNorm parameters while maintaining sparsity patterns through a fixed binary mask $M$.
\end{enumerate}

\subsection{SmoothQuant Adaptation} \label{sq_adapt}
While SmoothQuant~\cite{xiao2023smoothquant} was designed for quantization, we adapt it for sparsification by solving the inverse problem: rather than normalizing weights for discretization, we redistribute importance scores between activations and weights.

Given the weight matrix $W \in \mathbb{R}^{C_{\text{out}} \times C_{\text{in}}}$ and the input $x \in \mathbb{R}^{C_{\text{in}}}$, we compute channel-wise scaling factors $s_j = \max(|x_j|)/\max(|W_{:,j}|)$ for $j=1,...,C_{\text{in}}$. The diagonal scaling matrix $S = \text{diag}(s_1, \dots, s_{C_{\text{in}}})$ transforms:

\begin{equation}
\label{eq:sq}
W_{\text{ec}} = W \cdot S^{-1}, \quad x_{\text{scaled}} = x \cdot S
\end{equation}

\noindent This transformation preserves mathematical equivalence ($W_{\text{ec}}x_{\text{scaled}} = Wx$) while enabling clearer separation of salient vs. non-salient weights through magnitude redistribution.

\textbf{Implementation Note}: We only use $W_{\text{ec}}$ to calculate a metric to evaluate the importance of weights. The actual weights and model activations do not change.

\subsection{Variance Correction (VC)}
Motivated by \citeauthor{nagel2019data}'s bias correction for quantization, we propose variance correction to address distribution shifts from weight pruning:

\begin{equation}
\label{eq:vc}
W_{\neg\text{salient}}^{\text{(corrected)}} = W_{\neg\text{salient}} \cdot \sqrt \frac{\mathrm{Var}(W_{\text{dense}})}{\mathrm{Var}(W_{\neg\text{salient}}) + \epsilon}
\end{equation}

\noindent This rescaling preserves the original weight variance in pruned layers, maintaining stable activation statistics. Unlike bias correction, which requires learnable bias terms, our variance correction remains applicable to architectures without bias parameters.

\section{Models and Datasets}

Our study evaluates three language model families: LLaMA2 7B/13B  \cite{touvron2023llama}, LLaMA3 8B  \cite{grattafiori2024llama}, and Mistral 7B  \cite{jiang2023mistral7b}, which incorporate distinct architectural variations. While all models share foundational transformer-based designs, their implementations differ in attention mechanisms and optimization strategies, enabling comparative analysis of performance characteristics across model scales and design paradigms.

\textbf{For evaluation}, we calculate perplexity scores on the validation splits of two standard text corpora: WikiText \cite{merity2016pointer_wikitext} and C4 \cite{raffel2020exploring_c4}. To comprehensively assess reasoning capabilities, we employ five zero-shot benchmarks following established evaluation protocols: ARC Easy and Challenge \cite{clark2018think_arc} for structured knowledge reasoning, WinoGrande \cite{sakaguchi2021winogrande} for commonsense inference, HellaSwag \cite{zellers2019hellaswag} for contextual understanding, and PIQA \cite{bisk2020piqa} for physical interaction reasoning. The choice of baselines is similar to those in previous studies \cite{frantar2023sparsegpt, lee2024owq, frantar2022gptq, xiao2023smoothquant, zhelnin2024gift}. 

\input{results}

\section{Conclusion}
This work demonstrates that structured sparsity, when combined with targeted techniques for salient weight preservation, enables efficient LLM deployment without compromising performance. Our key findings are:

\begin{itemize}
    \item \textbf{8:16 Sparsity Provides Optimal Efficiency:} The 8:16 pattern delivers superior performance-efficiency tradeoffs. Our sparse LLaMa-2-13B matches the dense LLaMa-2-7B baseline while reducing computational overhead, validating its practical value for resource-constrained deployment.
    
    \item \textbf{Structured Sparsity for Salient Weights (\textsc{SSP for SW}) Outperforms Unstructured Approaches:} Recovering critical weights in structured patterns (4:256/8:256/16:256) consistently improves perplexity and accuracy across models. This approach yields better performance than unstructured sparsity while maintaining hardware efficiency.
    
    \item \textbf{Pre/Post-Processing Techniques:}
    \textit{SmoothQuant-Inspired Rebalancing} mitigates pruning-induced distribution shifts.
    Our novel \textit{Variance Correction} technique compensates for activation distribution changes.
    Combined with layer-wise \textit{EBFT} fine-tuning, these achieve the lowest perplexity (6.27 for Mistral with 8:16 sparsity).
    
    
\end{itemize}


\section{Limitations}
 The 8:16 sparsity pattern lacks current hardware support, however this work advocates for the development of such hardware desing in the future. Our evaluation is restricted to only three models (LLaMA-2/3, Mistral) on zero-shot reasoning benchmarks, limiting generalizability to smaller models, or tasks like instruction following or code generation.  Key ablation studies were omitted due to space constraints, obscuring the individual impact of components such as Variance Correction and SmoothQuant-inspired rebalancing.

 \section{Acknowledgement}

This work was supported by a grant for research center in the field of artificial intelligence, provided by the Ministry of Economic Development of the Russian Federation (agreement No. 139-15-2025-013, dated June 20, 2025, subsidy identifier 000000C313925P4B0002).


\bibliography{custom}

\clearpage
\input{appendix}

\end{document}

%% file: results.tex
\section{Results}
\label{sec:experiments}

\begin{table*}[!ht]
\centering
\resizebox{0.7\textwidth}{!}{%
  \renewcommand{\arraystretch}{1.0}
  \begin{tabular}{c|c|c|c|c|c|c}
    \toprule
    \textbf{Outliers:} & \multicolumn{2}{|c|}{\textbf{4:256}}& \multicolumn{2}{|c|}{\textbf{8:256}} & \multicolumn{2}{|c}{\textbf{16:256}}\\ \toprule
    \textbf{Sparsity:} & \textbf{2:4} & \textbf{8:16} & \textbf{2:4} & \textbf{8:16} & \textbf{2:4} & \textbf{8:16}\\ \hline
   \multicolumn{7}{c}{\textbf{C4}}    \\  \bottomrule    
    RIA+SQ & 55,97\% & \textbf{61,04}\% & 57,48\% & \textbf{61,18}\% & 59,16\% & \textbf{61,81}\% \\ 
    RIA+SQ+VC+EBFT & 57,87\% & \textbf{61,01}\% & 58,72\% & \textbf{60,95}\% & 59,74\% & \textbf{61,91}\% \\ 
    \toprule
    \multicolumn{7}{c}{\textbf{WikiText2}}    \\ 
    \bottomrule
   RIA+SQ & 56,44\% & \textbf{60,56}\% & 57,40\% & \textbf{61,12}\% & 59,41\% & \textbf{62,15}\% \\ 
    RIA+SQ+VC+EBFT & 57,84\% & \textbf{61,08}\% & 58,53\% & \textbf{61,27}\% & 60,20\% & \textbf{62,57}\% \\ 
    \bottomrule
  \end{tabular}
  }
  \caption{Mean accuracy for LLaMA2-7B with calibrations sets as C4 and WikiText2 on zero-shot tasks: ARC-c, ARC-e, PIQA, Winogrande, Hellaswag. Mean performance for dense model is 64.79\%.}
  \label{tab: salient_weights_zero_shot_tasks_llama2_7b}
\end{table*}

\begin{table*}[!ht]
\centering
\resizebox{0.7\textwidth}{!}{%
  \begin{tabular}{c|c|c|c|c|c|c}
    \toprule
   \textbf{Outliers:} & \multicolumn{2}{|c|}{\textbf{4:256}}& \multicolumn{2}{|c|}{\textbf{8:256}} & \multicolumn{2}{|c}{\textbf{16:256}}\\ \toprule
    \textbf{Sparsity:} & \textbf{2:4} & \textbf{8:16} & \textbf{2:4} & \textbf{8:16} & \textbf{2:4} & \textbf{8:16}\\ \hline
    \multicolumn{7}{c}{\textbf{C4}}    \\  \bottomrule    
    RIA+SQ & 60.18\% & 63.94\% & 61.19\% & 64.52\% & 62,96\% & 65,23\% \\ 
    RIA+SQ+VC+EBFT & 61.50\% & 64.52\% & 62.18\% & 64.16\% & 63,34\% & 65,13\% \\ 
    \toprule
    \multicolumn{7}{c}{\textbf{WikiText2}}    \\ 
    \bottomrule
    RIA+SQ & 60.28\% & 63.70\% & 60.92\% & 64.03\% & 63,03\% & 65,03\% \\ 
    RIA+SQ+VC+EBFT & 61.38\% & 63.63\% & 61.84\% & 63.76\% & 64,01\% & 65,40\% \\ 
    \bottomrule
  \end{tabular}
  }
  \caption{Mean accuracy for LLaMA2-13B with calibration sets as C4 and WikiText2 on zero-shot tasks: ARC-c, ARC-e, PIQA, Winogrande, Hellaswag. Mean performance for the dense model is 67.77\%.}
  \label{tab: salient_weights_zero_shot_tasks_llama2_13b}
\end{table*}


\begin{table}[!ht]
\centering
\resizebox{0.5\textwidth}{!}{%
\begin{tabular}{c|c|c|c}
\toprule
\textbf{Method} & \textbf{C4} & \textbf{WikiText2} & \textbf{Mean} \\ \hline
\textbf{Dense Model$^*$} & \textbf{5.47} & \textbf{5.47} & \textbf{5.47} \\
Magnitude$^*$ & 37.77 & 37.96 & 37.87 \\
RIA$^*$ & 11.27 & 10.91 & 11.09 \\
RIA +VC & 9.21 & 8.92 & 9.07 \\
RIA + SQ$^*$ & 10.71 & 10.22 & 10.47 \\
RIA + EBFT$^*$ & 8.73 & 8.47 & 8.60 \\
RIA + SQ + EBFT & 8.67 & 8.40 & 8.54 \\
RIA + SQ +VC+EBFT & 7.96 & 7.95 & 7.96 \\
\bottomrule
\end{tabular}
}
\caption{Perplexity results for LLaMA2-7B on C4 and WikiText2 datasets with 2:4 sparsity. The dense model achieves a perplexity of 5.47. Perplexity of RIA is improved by 28\% with all of the modifications applied. Baseline methods are specified with asterisks.}
\label{tab: ria_modifications_ppl}
\end{table}

\begin{table}[!ht]
\centering
\begin{tabular}{c|c|c}
\toprule
 \textbf{Outliers} & \textbf{LLaMA2-7B} & \textbf{LLaMA-13B} \\ \hline
 0\% & 37.96 & 18.46 \\
 1.56\% (4:256) & 23.06 & 14.59 \\
\bottomrule
\end{tabular}
\caption{Perplexity results for \textit{magnitude based pruning} of LLaMA2-7B and LLaMA-13B models, with  WikiText2 calibrration dataset, 2:4 sparsity pattrern is used.}
  \label{tab: salient_weights_ppl_magnitude_llama2_7b13b}
\end{table}

\begin{table*}[!ht]
\centering
\resizebox{0.7\textwidth}{!}{%
  \renewcommand{\arraystretch}{1.0}
  \begin{tabular}{c|c|c|c|c|c|c|c|c}
    \toprule
    \textbf{Outliers:} & \multicolumn{2}{|c|}{\textbf{-}} & \multicolumn{2}{|c|}{\textbf{4:256}}& \multicolumn{2}{|c|}{\textbf{8:256}} & \multicolumn{2}{|c}{\textbf{16:256}}\\ \toprule
    \textbf{Sparsity:} & \textbf{2:4} & \textbf{8:16} & \textbf{2:4} & \textbf{8:16} & \textbf{2:4} & \textbf{8:16} & \textbf{2:4} & \textbf{8:16}\\ \hline
   \multicolumn{7}{c}{\textbf{LLaMA3 (PPL=6.24)}}    \\  \bottomrule    
    RIA+SQ & 19.13 & 10.55 & 16.50 & 9.85 & 14.28 & 9.32 & 11.48 & 8.52 \\ 
    RIA+SQ+VC & 16.19 & 10.09 & 14.47 & 9.52 & 12.73 & 9.11 & 10.68 & 8.42 \\ 
    RIA+SQ+VC+EBFT & 10.49 & 8.08 & 9.18 & 7.88 & 9.53 & 7.69 & 8.24 & 7.40 \\ 
    \toprule
    \multicolumn{7}{c}{\textbf{Mistral (PPL=5.32)}}    \\ 
    \bottomrule
    RIA+SQ & 9.34 & 6.74 & 8.48 & 6.52 & 7.80 & 6.34 & 6.97 & 6.06 \\ 
    RIA+SQ+EBFT & 7.10 & 6.14 & 6.86 & 6.06 & 6.59 & 5.96 & 6.27 & 5.82 \\ 
    \bottomrule
  \end{tabular}
  }
  \caption{Perplexity for \textbf{LLaMA3-8B} and \textbf{Mistral} with WikiText2 as calibration set on different sparsity and outliers type.}
  \label{tab: salient_weights_ppl_llama3_mistral}
\end{table*}

This section presents the results of computational experiments.
(1)~We analyze the effectiveness of SmoothQuant (SQ) adaptation and Variance Correction (VC).
(2)~We evaluate a semi-structured 4:256 sparsity pattern for storing weight outliers.
(3)~We compare weight sparsification with 2:4 and 8:16 patterns, both with and without fine-tuning to demonstrate effectivness of 8:16 pattern.
(4)~Finally, we contrast structured and unstructured sparsity to validate semi-structured patterns for outlier storage.

\textbf{\numcircledtikz{1} Variance Correction and Weight Equalization with SmoothQuant.}
To evaluate VC and SQ, we measured perplexity in the pruned LLaMA2-7B model (Table \ref{tab: ria_modifications_ppl}). Both methods reduced perplexity versus standard RIA pruning on C4 and Wikitext2 datasets, with VC being more efficeint.  However, layer-wise fine-tuning via EBFT yielded the most significant improvement, outperforming both VC and SQ. Notably, all methods independently enhanced performance, and their combination achieved the lowest perplexity across datasets.

\textbf{\numcircledtikz{2} Structured Outlier Storage for RIA and Magnitude-Based Sparsification.}
We recovered salient weights in semi-structured 2:4-pruned linear layers of LLaMA2-7B/13B models, storing outliers in 4:256 patterns. As Table~\ref{tab: salient_weights_ppl_magnitude_llama2_7b13b} shows, preserving outliers—even under magnitude-based pruning—improves model performance. This section focuses on magnitude-based approaches to isolate implicit effects of other pruning techniques.
It is evident that recovering even a small number of salient weights using the 4:256 format significantly improves perplexity for both models. Thus, structural recovery of salient weights can be integrated with various pruning methods to enhance the generative capabilities of pruned models.

\textbf{ \numcircledtikz{3} Comparison of 2:4 and 8:16 structured patterns.} 
Here, we compare perforomance for all models on reasoning tasks  for two sparsity patterns 2:4 and 8:16. 

We perform structural recovery of salient weights in the linear layers of language models subjected to semi-structured 2:4 and 8:16 pruning patterns. For each linear layer, salient weights are extracted in structural pruning patterns of 4:256, 8:256, and 16:256 and stored in a separate matrix.

\textbf{LLaMA2-7B and LLaMA2-13B.}
Tables \ref{tab: salient_weights_zero_shot_tasks_llama2_7b} and \ref{tab: salient_weights_zero_shot_tasks_llama2_13b} present the mean accuracy across five zero-shot tasks for LLaMA2-7B and LLaMA2-13B. The results demonstrate that increasing the number of salient weights consistently improves model accuracy for both the RIA+SQ and RIA+SQ+VC+EBFT pruning methods when using the wikitext2 dataset. However, a slight decrease in accuracy is observed for RIA+SQ+VC+EBFT on the c4 dataset. The best performance for pruned models is achieved when isolating salient weights in the 16:256 pattern. 

\textbf{LLaMA3.1-8B and Mistral.}
Since wikitext2 yields better results with an increasing number of salient weights, we select this dataset for calibrating more modern models, such as LLaMA3.1-8B and Mistral, during pruning. Table \ref{tab: salient_weights_ppl_llama3_mistral} shows the perplexity on wikitext2 for \textbf{LLaMA3.1-8B} and \textbf{Mistral} under 2:4 and 8:16 pruning patterns, depending on the number of structural salient weights. For Mistral, we omitted VC as it significantly degrades generative capabilities. In contrast, VC substantially improves perplexity for LLaMA3.1-8B.
Mistral demonstrates superior robustness to pruning compared to LLaMA3. Under the 2:4 pruning pattern with the RIA+SQ method, LLaMA3's perplexity increases significantly, by up to 3.07× (from 6.24 to 19.13), whereas Mistral's perplexity rises by only 1.75× (from 5.32 to 9.34). For the 8:16 pruning pattern, the increase in perplexity is more moderate, with Mistral experiencing a 1.27× increase and LLaMA3 a 1.69× increase. Fine-tuning with EBFT enhances both models, regardless of the pruning pattern. Specifically, for LLaMA3, the increase in perplexity is limited to 1.68× (from 6.24 to 10.49), and for Mistral, it is reduced to 1.33× (from 5.32 to 7.10).

\textbf{The incorporation of structural salient weights in the 2:4 and 8:16 pruning patterns significantly improves perplexity.} We can observe this for both models. Increasing the number of recovered salient weights allows for a substantial reduction in perplexity following pruning. The models achieve optimal perplexity metrics when employing a 16:256 recovery scheme in conjunction with EBFT. In this scenario, LLaMA3's perplexity metrics are 8.24 and 7.40, while Mistral exhibits perplexity values of 6.27 and 5.82, depending on the pruning pattern used (2:4 or 8:16).

These conclusions are further supported by the accuracy results of LLaMA3 and Mistral on zero-shot tasks. The table demonstrates that combining pruning with EBFT on Wikitext-2, and salient weight recovery enhances model performance. For the 16:256 recovery scheme, LLaMA3 shows a minimal accuracy drop of 7.48\% and 4.34\% (from 69.23 to 61.75 and from 69.23 to 64.89), while Mistral experiences drops of 6.43\% and 3.28\% (from 68.55 to 62.11 and from 68.55 to 65.27), depending on the pruning pattern used (2:4 or 8:16).

\textbf{ Structured VS unstructured salient weights.}
To emphasize the effectiveness of recovering salient weights in a structural format, we compare the performance of models subjected to pruning using RIA, SQ, VC, and EBFT methods, alongside both structural and unstructural salient weight recovery. For this comparison, the number of salient weights recovered in the unistructural format is kept comparable to that of the structural format, with EBFT conducted on Wikitext-2. The results are a presnted in Table~\ref{tab:llama_sparsity_comparison}. We see that both structural and unstructured weight recovery enhance model performance following pruning. However, the use of a structured format consistently yields better metrics compared to unstructured salient weights. Therefore, structural salient weights not only facilitate more efficient model inference but also contribute to improved model performance after pruning compared to their unstructured counterparts.


%% file: appendix.tex
\newpage
\appendix
\section*{Appendix}

This appendix presents additional experimental results comparing different sparsity patterns and quantization-aware fine-tuning strategies across several large language models.

Table~\ref{tab:llama_sparsity_comparison} reports the zero-shot accuracy of LLaMA-2-7B and LLaMA-2-13B under varying levels of sparsity specifically, 4/256 ($\approx$0.0156\%), 8/256 ($\approx$0.0313\%), and 16/256 ($\approx$0.0625\%) using both unstructured and semi-structured pruning approaches. All models are calibrated on the WikiText dataset. Results indicate that semi-structured sparsity consistently matches or slightly outperforms unstructured sparsity across both model sizes and sparsity levels, suggesting that structured pruning can preserve task performance while offering hardware-friendly compression benefits.

Table~\ref{tab: salient_weights_zero_shot_tasks_llama3_mistral} evaluates the impact of progressively incorporating advanced compression techniques including salient weight identification (RIA), smooth quantization (SQ), variance-aware calibration (VC), and efficient bias fine-tuning (EBFT) on zero-shot performance for LLaMA3-8B and Mistral. Accuracy is averaged across five standard benchmarks: ARC-c, ARC-e, PIQA, Winogrande, and Hellaswag, with WikiText2 used for calibration. The baseline accuracies (without compression) are 69.23\% for LLaMA3-8B and 68.55\% for Mistral. The results demonstrate that combining RIA+SQ with VC and EBFT yields consistent improvements over simpler pipelines, particularly at higher sparsity levels (e.g., 16:256 outliers with 2:4 or 8:16 block sparsity). Notably, Mistral shows strong resilience to compression, maintaining performance close to its uncompressed baseline even under aggressive sparsification.
\begin{table}[ht]
\centering
\resizebox{0.5\textwidth}{!}{%
  \renewcommand{\arraystretch}{1.0}
  \begin{tabular}{c|c|c|c|c|c|c}
    \toprule
    \textbf{Sparsity (\%):} & \multicolumn{2}{|c|}{\textbf{4/256} ($\approx 0.0156\%$)} & \multicolumn{2}{|c|}{\textbf{8/256} ($\approx 0.0313\%$)} & \multicolumn{2}{|c}{\textbf{16/256} ($\approx 0.0625\%$)} \\ \toprule

    \multicolumn{7}{c}{\textbf{LLaMA-2-7B}}    \\  \bottomrule    
    Unstructured & 56.43\% & 60.08\% & 57.55\% & 60.55\% & 59.09\% & 61.88\% \\ 
    Semi-structured & \textbf{57,84}\% & \textbf{61,08}\% & \textbf{58,53}\% & \textbf{61,27}\% & \textbf{60,20}\% & \textbf{62,57}\% \\ 
    \toprule
    \multicolumn{7}{c}{\textbf{LLaMA-2-13B}}    \\ 
    \bottomrule
    Unstructured & 61.37\% & 62.94\% & \textbf{62.03}\% & 63.48\% & 62.796\% & 64.20\% \\ 
    Semi-structured & \textbf{61.38}\% & \textbf{63.63}\% & 61.84\% & \textbf{63.76}\% & \textbf{64,01}\% & \textbf{65,40}\% \\ 
    \bottomrule
  \end{tabular}
  }
  \caption{Comparision between unstructured and semi structured sparsity patterns. Accuracy is reported for LLaMA-2 models under varying sparsity levels correspoining to  following sparsity patterns: 4/256, 8/256/ and 16/266. WikiText is used for calibration.}
  \label{tab:llama_sparsity_comparison}
\end{table}

\begin{table}[ht]
\centering
\resizebox{0.5\textwidth}{!}{%
  \renewcommand{\arraystretch}{1.0}
  \begin{tabular}{c|c|c|c|c|c|c|c|c}
    \toprule
    \textbf{Outliers:} & \multicolumn{2}{|c|}{\textbf{-}} & \multicolumn{2}{|c|}{\textbf{4:256}}& \multicolumn{2}{|c|}{\textbf{8:256}} & \multicolumn{2}{|c}{\textbf{16:256}}\\ \toprule
    \textbf{Sparsity:} & \textbf{2:4} & \textbf{8:16} & \textbf{2:4} & \textbf{8:16} & \textbf{2:4} & \textbf{8:16} & \textbf{2:4} & \textbf{8:16}\\ \hline
   \multicolumn{7}{c}{\textbf{LLaMA3 (acc=69.23\%)}}    \\  \bottomrule    
    RIA+SQ & 51.16\% & 59.71\% & 52.85\% & 60.60\% & 54.75\% & 61.81\% & 58.01\% & 63.75\% \\ 
    RIA+SQ+VC & 52.85\% & 60.19\% & 54.25\% & 60.98\% & 55.97\% & 62.29\% & 58.87\% & 63.98\% \\ 
    RIA+SQ+VC+EBFT & 55.78\% & 62.01\% & 58.41\% & 62.71\% & 58.95\% & 63.20\% & 61.75\% & 64.89\% \\ 
    \toprule
    \multicolumn{7}{c}{\textbf{Mistral (acc=68.55\%)}}    \\ 
    \bottomrule
    RIA+SQ & 56.72\% & 63.04\% & 58.30\% & 63.81\% & 59.12\% & 64.09\% & 61.38\% & 64.87\% \\ 
    RIA+SQ+EBFT & 58.49\% & 63.63\% & 59.26\% & 64.07\% & 60.68\% & 64.48\% & 62.11\% & 65.27\% \\ 
    \bottomrule
  \end{tabular}
  }
  \caption{Mean accuracy for \textbf{LLaMA3-8B} and \textbf{Mistral} with WikiText2 as calibration set on zero-shot tasks: ARC-c, ARC-e, PIQA, Winogrande, Hellaswag.}
  \label{tab: salient_weights_zero_shot_tasks_llama3_mistral}
\end{table}